\documentclass{article}

\usepackage[final]{neurips_2022}

\usepackage[utf8]{inputenc} 
\usepackage[T1]{fontenc}    
\usepackage{hyperref}       
\usepackage{url}            
\usepackage{booktabs}       
\usepackage{amsfonts}       
\usepackage{nicefrac}       
\usepackage{microtype}      
\usepackage{xcolor}         
\usepackage{graphicx}
\usepackage[inline]{enumitem}
\usepackage{subcaption}
\usepackage{multirow}
\usepackage{bold-extra}
\usepackage{courier}

\newcommand{\pytail}[0]{\texttt{PyTAIL}}

\title{\pytail{}: Interactive and Incremental Learning of NLP Models with Human in the Loop for Online Data}

\author{%
  Shubhanshu Mishra\thanks{Work done while at University of Illinois at Urbana-Champaign.} \\
  shubhanshu.com\\
  \texttt{mishra@shubhanshu.com} \\
  \And
  Jana Diesner \\
  University of Illinois at Urbana-Champaign \\
  \texttt{jdiesner@illinois.edu}
}

\begin{document}

\maketitle

\begin{abstract}
Online data streams make training machine learning models hard because of distribution shift and new patterns emerging over time. For natural language processing (NLP) tasks that utilize a collection of features based on lexicons and rules, it is important to adapt these features to the changing data. To address this challenge we introduce \pytail{}, a python library, which allows a human in the loop approach to actively train NLP models. \pytail{} enhances generic active learning, which only suggests new instances to label by also suggesting new features like rules and lexicons to label. Furthermore, \pytail{} is flexible enough for users to accept, reject, or update rules and lexicons as the model is being trained. Finally, we simulate the performance of \pytail{} on existing social media benchmark datasets for text classification. We compare various active learning strategies on these benchmarks. The model closes the gap with as few as 10\% of the training data. Finally, we also highlight the importance of tracking evaluation metric on remaining data (which is not yet merged with active learning) alongside the test dataset. This highlights the effectiveness of the model in accurately annotating the remaining dataset, which is especially suitable for batch processing of large unlabelled corpora. \pytail{} will be available at \url{https://github.com/socialmediaie/pytail}.

\end{abstract}

\section{Introduction}

Analysis of large scale natural language corpora often requires annotation of dataset in a given domain with pre-trained models. Generally, these models are pre-trained on a fixed training dataset which is often different from the domain of the dataset under consideration. This often leads to poor performance of the model on this new domain. One way to address this gap is to utilize domain adaptation \citep{Sarawagi2007,Daume2007} to improve the model accuracy. However, efficient domain adaptation requires labeled training data from the new domain, which is costly to acquire. The problem gets compounded for social media data, for which the vocabulary and language usage continuously evolve over time.
Take the example of sentiment classification, where the ways of expressing the same opinion also change with time. For example, the opinion label of the phrase ``you are just like \textit{subject}", will depend on the general opinion about \textit{``subject"} when the phrase was expressed. Similarly, many new words are coined on social media \citep{Eisenstein2013,Gupta2010}. This poses a challenge for maintaining these models retain their accuracy over time.
In this work, we propose an approach to alleviate this issue by creating a system based on active human-in-the-loop learning which incrementally updates an existing classifier by requiring an user to provide few new examples from the new data. Traditionally, this setup, called active learning \citep{Settles2009} only deals with suggesting new training examples to annotate. However, since many NLP models use feature based on existing rules or lexicons, with changing data characteristics it may be more desirable to also suggest rule and lexicon updates in the model. Our system \pytail{} (Python Text Analysis and Incremental Learning) addresses the issues highlighted here by allowing human-in-the-loop active learning systems to integrate new data points, rules, and lexicons. Our main contributions are as follows:
\begin{enumerate*}[label=(\roman*)]
    \item Introduce \pytail{}, an open source tool with an active learning workflow which uses new data, rules, and lexicons to continuously train NLP models. 
    \item Introduce a social media text classification benchmark for active learning research.
    \item Introduce an evaluation setup on unconsumed data in active learning to quantify how quickly a corpus can be fully annotated with a reasonable accuracy.
\end{enumerate*}

\section{Incremental learning of models with human in the loop}

\begin{figure}
\centering
\includegraphics[width=\linewidth]{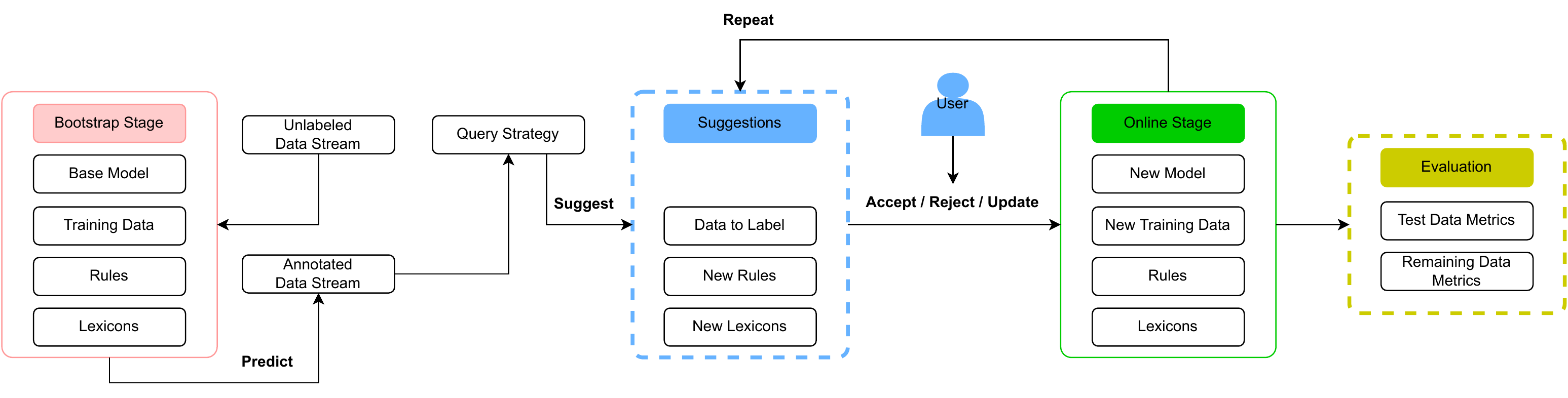}
\caption{\textbf{\pytail{} Workflow}: Given a user and an unlabeled data stream, along with some bootstrapping artifacts, \pytail{} suggests data instances, rules, and lexicons which can be merged with bootstrapping artifacts to continuously create new model.}
\label{fig:pytail-workflow}
\end{figure}

In this section we describe \pytail{} (Python Text Analysis and Incremental Learning). \pytail{}'s goal is to enable efficient construction of training data using active learning, while supporting incremental learning of models using the most recent data. A description of \pytail{} workflow is shown in figure \ref{fig:pytail-workflow}. \pytail{} is built with the following features in mind: 
\begin{enumerate*}[label=(\roman*)]
\item Low cost of continuous training data acquisition
\item Incorporation of domain knowledge using lexicon and rules
\item Efficient update of model using only the newly acquired training data.
\end{enumerate*}

\paragraph{Overview} As shown in figure \ref{fig:pytail-workflow}, the user starts with a collection of artifacts in the Bootstrap Stage. This can include an pre-trained model, a small seed training dataset, existing rules, and lexicons. Next, the user introduces their unlabeled data stream from their domain of interest, e.g. social media corpora. The bootstrap artifacts are used to predict this data stream. These predictions are then fed to the query strategy (described below) to identify artifacts for the suggestion stage. The user can then accept, reject, update these suggestions or even introduce new suggestions. Next, the model is updated using updated artifacts such that the rules and lexicons are used for updating the model features and the annotated data is used for updating the model. Finally, \pytail{} shows continuous evaluation metrics which include metric on a test set, user accepted training set, and unobserved data stream. This process is repeated till a stopping criteria is met, e.g. the exhaustion of data stream or achieving reasonable evaluation score. \pytail{} supports two modes for training, one is human in the loop (HITL) mode, and another is simulation mode. The simulation model uses pre-defined heuristics to simulate human actions based on model prediction scores. The default model when applied to benchmark datasets is the simulation mode.

\paragraph{Human in the loop (HITL) mode} In the HITL mode, \pytail{} uses the pre-trained model to suggest top $K$ instances to the user. The user can sort the instances using the scoring criterion. In order to reduce the cognitive work of labeling an instance from scratch, the user is shown the model predictions (as well as the label probability). The user is only required to edit the labels if they disagree. Model predictions for all the unlabeled instances from the top suggestions are now used as gold labels and fed to the model during the update process (this is similar to self-supervision with the possibility of human intervention). The user is also shown the prominent features for that instance, the user can select these features and mark them as useful or useless. Lexicon matches with the annotations are also shown, along with prominent key phrases in the unlabeled data stream. The user can choose to update the lexicon with these new suggestions. Once the model update has happened, the user is provided feedback on the change in model evaluation on a held out data.

\section{Benchmark for social media active learning}
We introduce an active learning benchmark of 10 social media text classification datasets consisting of 200K posts. These datasets cover sentiment classification, abusive content identification, and uncertainty indication.

\setlength{\tabcolsep}{2pt}
\begin{table}[!htbp]
    \begin{subfigure}[t]{0.5\linewidth}
    \centering
    \caption[Description of sentiment classification datasets]{Description of sentiment classification datasets. Datasets clustered together are enclosed between horizontal lines. Labels are \textit{negative}, \textit{neutral}, \textit{positive}.}
    \label{tab:mtl_sentiment_data}
    \begin{tabular}{llrrr}
    \toprule
    \textbf{data} & \textbf{split} & tokens &  tweets &  vocab\\
    \midrule
    \multirow{3}{*}{\textbf{Airline}} & \textbf{dev} &   20079 &     981 &   3273 \\
            & \textbf{test} &   50777 &    2452 &   5630 \\
            & \textbf{train} &  182040 &    8825 &  11697 \\
    \cline{1-5}
    \multirow{3}{*}{\textbf{Clarin}} & \textbf{dev} &   80672 &    4934 &  15387 \\
            & \textbf{test} &  205126 &   12334 &  31373 \\
            & \textbf{train} &  732743 &   44399 &  84279 \\
    \cline{1-5}
    \multirow{3}{*}{\textbf{GOP}} & \textbf{dev} &   16339 &     803 &   3610 \\
            & \textbf{test} &   41226 &    2006 &   6541 \\
            & \textbf{train} &  148358 &    7221 &  14342 \\
    \cline{1-5}
    \multirow{3}{*}{\textbf{Healthcare}} & \textbf{dev} &   15797 &     724 &   3304 \\
            & \textbf{test} &   16022 &     717 &   3471 \\
            & \textbf{train} &   14923 &     690 &   3511 \\
    \cline{1-5}
    \multirow{3}{*}{\textbf{Obama}} & \textbf{dev} &    3472 &     209 &   1118 \\
            & \textbf{test} &    8816 &     522 &   2043 \\
            & \textbf{train} &   31074 &    1877 &   4349 \\
    \cline{1-5}
    \multirow{3}{*}{\textbf{SemEval}} & \textbf{dev} &  105108 &    4583 &  14468 \\
            & \textbf{test} &  528234 &   23103 &  43812 \\
            & \textbf{train} &  281468 &   12245 &  29673 \\
    \bottomrule
    \end{tabular}
    \end{subfigure}
    \hfill
\begin{subfigure}[t]{0.47\linewidth}
\begin{subfigure}[t]{1\linewidth}
\centering
    \caption[Description of abusive content classification datasets]{Description of abusive content classification datasets. Datasets which are clustered together are enclosed between horizontal lines. Labels for Founta are \textit{abusive}, \textit{hateful}, \textit{normal}, and \textit{spam}. Labels for WaseemSRW are \textit{none}, \textit{racism}, and \textit{sexism}. }
    \label{tab:mtl_abusive_data}
    \begin{tabular}{llrrr}
    \toprule
    \textbf{data} & \textbf{split} & tokens &  tweets &  vocab\\
    \midrule
    \multirow{3}{*}{\textbf{Founta}} & \textbf{dev} &  102534 &    4663 &   22529 \\
          & \textbf{test} &  256569 &   11657 &   44540 \\
          & \textbf{train} &  922028 &   41961 &  118349 \\
    \cline{1-5}
    \multirow{3}{*}{\textbf{WaseemSRW}} & \textbf{dev} &   25588 &    1464 &    5907 \\
              & \textbf{test} &   64893 &    3659 &   10646 \\
              & \textbf{train} &  234550 &   13172 &   23042 \\
    \bottomrule
    \end{tabular}
\end{subfigure}
\begin{subfigure}[t]{1\linewidth}
\centering
    \caption[Description of uncertainty indicators dataset stats]{Description of uncertainty indicators dataset. Datasets which are clustered together are enclosed between horizontal lines. Labels for Riloff are \textit{sarcasm} and \textit{not sarcasm}. Labels are for Swamy are \textit{definitely no}, \textit{definitely yes}, \textit{probably no}, \textit{probably yes}, and \textit{uncertain}.}
    \label{tab:mtl_uncertainty_data}
    \begin{tabular}{llrrr}
    \toprule
    \textbf{data} & \textbf{split} & tokens &  tweets &  vocab\\
    \midrule
    \multirow{3}{*}{\textbf{Riloff}} & \textbf{dev} &    2126 &     145 &   1002 \\
      & \textbf{test} &    5576 &     362 &   1986 \\
      & \textbf{train} &   19652 &    1301 &   5090 \\
    \cline{1-5}
    \multirow{3}{*}{\textbf{Swamy}} & \textbf{dev} &    1597 &      73 &    738 \\
          & \textbf{test} &    3909 &     183 &   1259 \\
          & \textbf{train} &   14026 &     655 &   2921 \\
    \bottomrule
    \end{tabular}
\end{subfigure}
\end{subfigure}
    \caption{Benchmark Datasets for Social Media Active Learning}
    \label{tab:pytail-benchmark-data}
\end{table}

\subsection{Sentiment classification}
For sentiment classification we use the same data as in \citep{Mishra2018}. A description of these data is shown in table \ref{tab:mtl_sentiment_data}. Clarin \cite{Mozetic2016} and SemEval are the two largest corpora. However, SemEval has a larger test set. All the sentiment datasets use the traditional labels of positive, neutral, and negative for labeling the tweets.
\subsection{Abusive content classification}
The second task we consider is abusive content classification. This task has recently gained prominence, owing to the the growth of abusive content on social media platforms. We utilize two datasets of abusive content. The first data is Founta from \cite{Founta2018}, which tags tweets as \textit{abusive, hateful, normal, spam}. The second dataset is WaseemSRW from \cite{Waseem2016}. It tags the data as \textit{none, racism, sexism}. The rationale for including both these data under the same task it the core idea of identifing abusive content either direct or using racist or sexist variation. A description of these data is shown in table \ref{tab:mtl_abusive_data}. 
\subsection{Uncertainty indicators}
Finally, we consider a collection of datasets for the task of identifying uncertainty indicators. Uncertainty indicators are defined as indicators in text which capture a level of uncertainty about the text, e.g., veridictality or sarcasm (uncertainty in intended meaning). We consider two datasets for this task as well. The first dataset is Riloff from \cite{Riloff2013SarcasmSituation}. This dataset consists of tweets annotated for sarcasm and non-sarcasm. The second dataset is Swamy from \cite{Swamy2017}. This dataset tries to identify the level of veridictality or degree of belief expressed in the tweet. The label set for this data is \textit{definitely no, probably no, uncertain, probably yes, definitely yes}. A description of these data is shown in \ref{tab:mtl_uncertainty_data}.

\section{\pytail{} for Social Media Text Classification}

\paragraph{Model} We use a logistic regression model with $L_2$ regularization. The regularization parameter is selected for each model using cross validation. We track the model scores on the held out test as well as validation data. Each text is represented using a set of features. Each tweet is tokenized and pre-processed by normalizing all mentions of hashtags, URLs, and mentions. We also use a large sentiment lexicon\footnote{https://github.com/juliasilge/tidytext/blob/master/data-raw/sentiments.csv}. Furthermore, we suggest including a domain specific negative filter, i.e., words which should not be used to identify classification signals. For sentiment classification this can be entities in the corpora which should not bias the model.

\paragraph{Query selection strategies}
Active learning algorithms \citep{Settles2009} identify most informative instances from unlabeled data that can be used to construct a high quality training dataset. The process of identifying informative instances is called \textbf{query selection}. Top instances $X_{selected}$ from the unlabeled data $X_{unlabeled}$ are identified based on a score. We consider two types of score:
\begin{enumerate*}[label=(\roman*)]
\item $entropy = \sum_i p_i*log(p_i)$ - higher is better
\item $min\mathit{-}margin = \max_{i \ne \star} \{p_i - p_{\star} \mid p_{\star} = \max_j p_j\}$ - lower is better.
\end{enumerate*}
The entropy based scoring favors model predictions with highest randomness. The min-margin based scoring is useful in ensuring that the difference between the top prediction score and the second top prediction score is less. The selection is done using three strategies: 
\begin{enumerate*}[label=(\roman*)]
\item \textbf{$Rand$}: Instances are selected randomly without considering their scores, this acts as a baseline. 
\item \textbf{$X_{top}$}: Top $K$ instances are selected based on their scores ($X$). 
\item \textbf{$X_{prop}$}: $K$ instances are sampled proportional to their scores ($X$). This adds a degree of randomness to the top k strategy. 
\end{enumerate*}
These new instances are then added to the existing training instances $X_{train} = X_{train} \cup X_{selected}$, and the model is retrained. 

\setlength{\tabcolsep}{2pt}
\begin{table}[!htbp]
    \centering
    \caption{Performance of query strategies across datasets using around 10\% training dataset.}
    \label{tab:10pct-results}
    \begin{tabular}{llrrrrr|rrrrr}
\toprule
             task & dataset &  round &   $N$ & $N_{left}$ &  $\%_{used}$ &  Full &  Rand &  $E_{top}$ &  $E_{prop}$ &  $M_{top}$ &  $M_{prop}$ \\
\midrule
\multicolumn{12}{c}{Test Dataset}\\
\midrule
ABUSIVE & Founta &     42 &  41,861 &   37,661 &      0.10 &  0.79 &  0.77 &     0.78 &      0.78 &     0.79 &      0.77 \\
             & WaseemSRW &     14 &  13,072 &   11,672 &      0.11 &  0.82 &  0.79 &     0.78 &      0.77 &     0.78 &      0.76 \\
SENTIMENT & Airline &      9 &   8,725 &    7,825 &      0.10 &  0.82 &  0.76 &     0.78 &      0.79 &     0.77 &      0.77 \\
             & Clarin &     45 &  44,299 &   39,799 &      0.10 &  0.66 &  0.63 &     0.61 &      0.62 &     0.63 &      0.63 \\
             & GOP &      8 &   7,121 &    6,321 &      0.11 &  0.67 &  0.63 &     0.64 &      0.63 &     0.62 &      0.64 \\
             & Healthcare &      1 &     590 &      490 &      0.17 &  0.59 &  0.64 &     0.60 &      0.61 &     0.60 &      0.60 \\
             & Obama &      2 &   1,777 &    1,577 &      0.11 &  0.63 &  0.56 &     0.60 &      0.58 &     0.59 &      0.57 \\
             & SemEval &     13 &  12,145 &   10,845 &      0.11 &  0.65 &  0.59 &     0.60 &      0.61 &     0.58 &      0.61 \\
UNCERTAINITY & Riloff &      2 &   1,201 &    1,001 &      0.17 &  0.78 &  0.77 &     0.76 &      0.77 &     0.76 &      0.79 \\
             & Swamy &      1 &     555 &      455 &      0.18 &  0.39 &  0.39 &     0.40 &      0.39 &     0.34 &      0.31 \\
\midrule
\multicolumn{12}{c}{Remaining Dataset}\\
\midrule
ABUSIVE & Founta &     42 &  41,861 &   37,661 &      0.10 &   NaN &  0.77 &     0.80 &      0.78 &     0.81 &      0.78 \\
             & WaseemSRW &     14 &  13,072 &   11,672 &      0.11 &   NaN &  0.78 &     0.79 &      0.77 &     0.80 &      0.76 \\
SENTIMENT & Airline &      9 &   8,725 &    7,825 &      0.10 &   NaN &  0.75 &     0.79 &      0.79 &     0.80 &      0.78 \\
             & Clarin &     45 &  44,299 &   39,799 &      0.10 &   NaN &  0.62 &     0.62 &      0.62 &     0.64 &      0.63 \\
             & GOP &      8 &   7,121 &    6,321 &      0.11 &   NaN &  0.62 &     0.64 &      0.62 &     0.63 &      0.63 \\
             & Healthcare &      1 &     590 &      490 &      0.17 &   NaN &  0.53 &     0.56 &      0.53 &     0.47 &      0.50 \\
             & Obama &      2 &   1,777 &    1,577 &      0.11 &   NaN &  0.54 &     0.56 &      0.57 &     0.56 &      0.56 \\
             & SemEval &     13 &  12,145 &   10,845 &      0.11 &   NaN &  0.61 &     0.62 &      0.62 &     0.63 &      0.62 \\
UNCERTAINITY & Riloff &      2 &   1,201 &    1,001 &      0.17 &   NaN &  0.80 &     0.82 &      0.84 &     0.82 &      0.81 \\
             & Swamy &      1 &     555 &      455 &      0.18 &   NaN &  0.37 &     0.40 &      0.40 &     0.33 &      0.36 \\
\bottomrule
\end{tabular}
\end{table}

\begin{figure}
    \centering
     \begin{subfigure}[b]{0.95\textwidth}
         \centering
         \includegraphics[width=\textwidth]{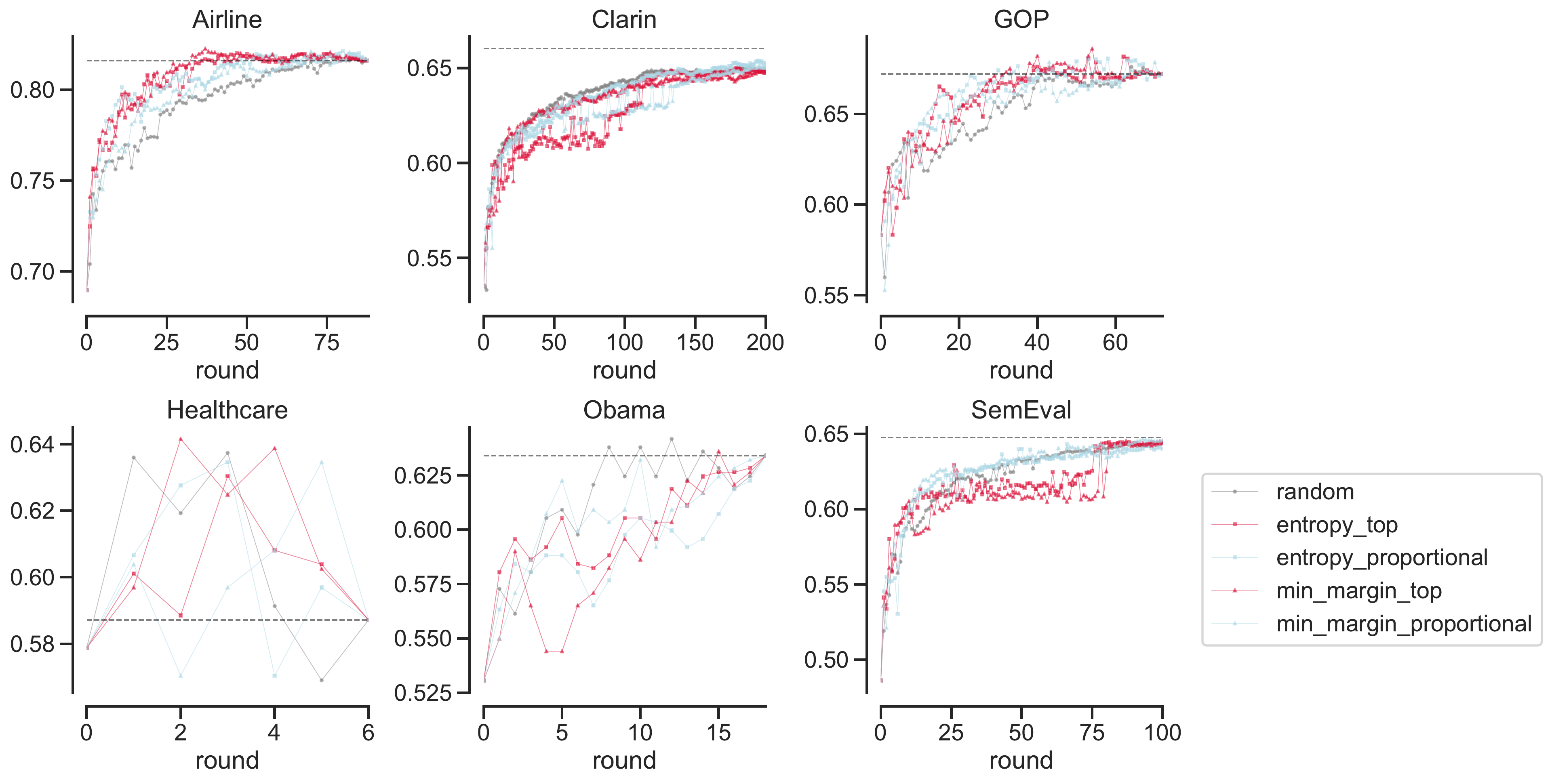}
         \caption{Sentiment classification}
         \label{fig:hitl_sentiment_test_results}
     \end{subfigure}
     \begin{subfigure}[b]{0.95\textwidth}
         \centering
         \includegraphics[width=\textwidth]{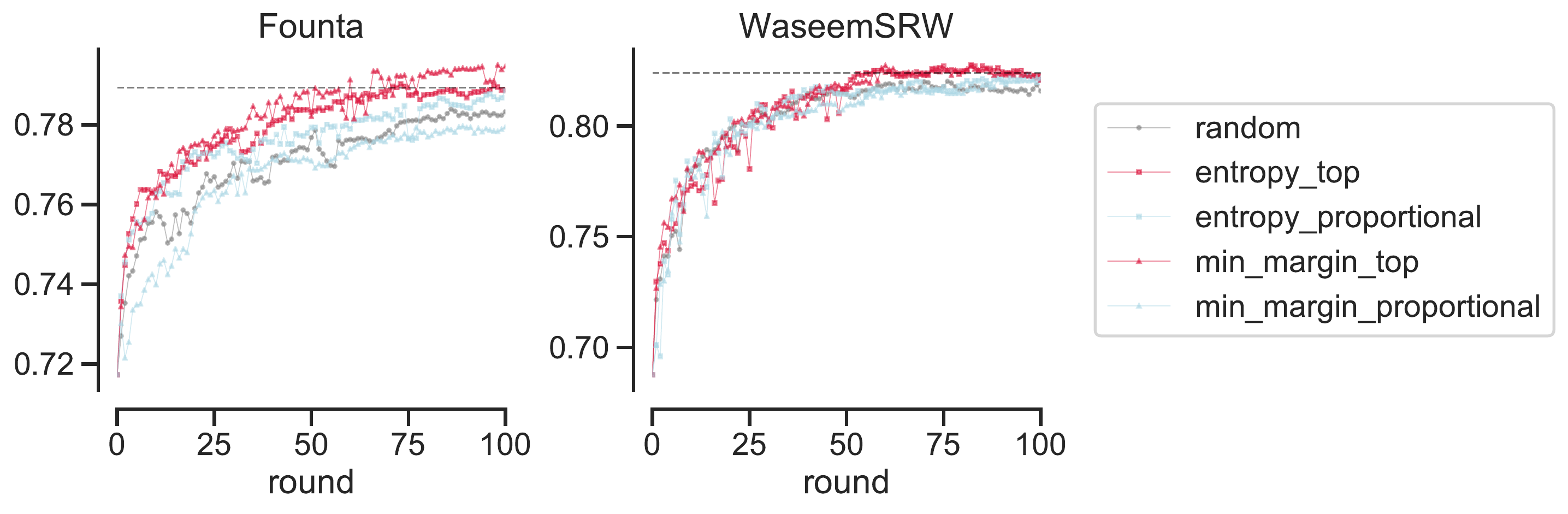}
         \caption{Abusive content detection}
         \label{fig:hitl_abusive_test_results}
     \end{subfigure}
     \begin{subfigure}[b]{0.95\textwidth}
         \centering
         \includegraphics[width=\textwidth]{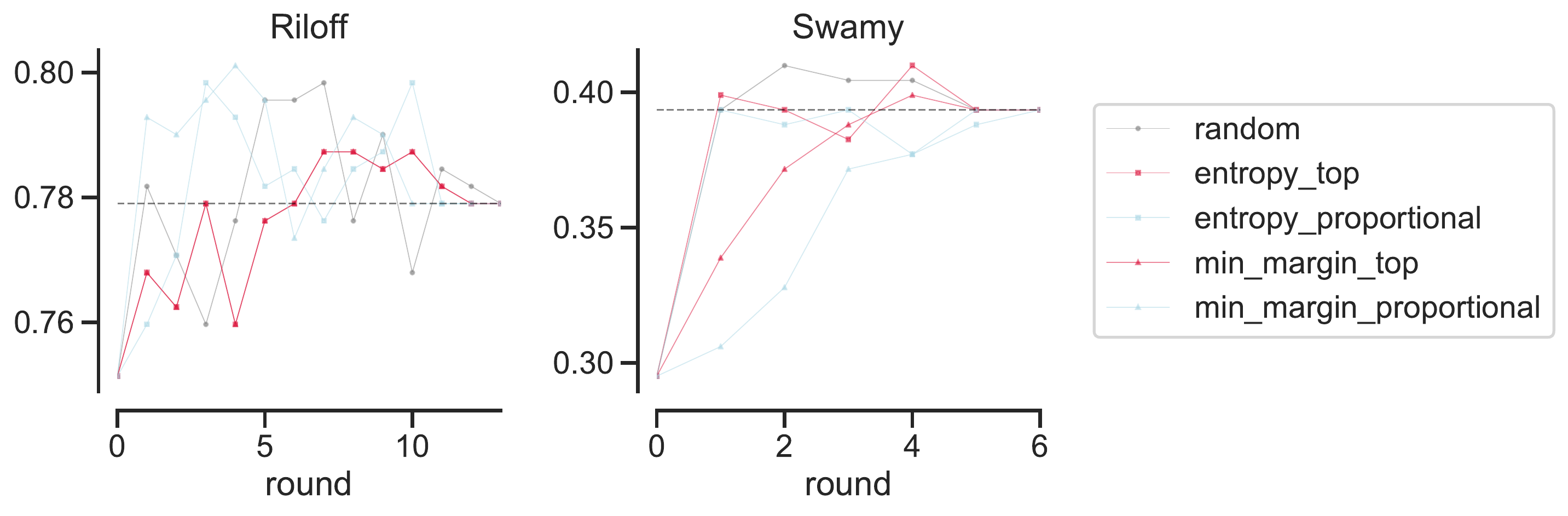}
         \caption{Uncertainty indicators}
         \label{fig:hitl_uncertainty_test_results}
     \end{subfigure}
    \caption[Active learning performance on multiple classification tasks]{Progression of active learning classifier performance (micro f1-score) on the respective test set across 100 rounds of active learning (200 for Clarin). The annotation budget for each round is 100 instances, and the model is warm started with 100 random samples of the training data. Black dotted line is the classifier performance when trained on all of the training data. Data ordered alphabetically and X and Y axes are not shared.}
    \label{fig:hitl_all_classification_test_results}
\end{figure}

\begin{figure}
    \centering
     \begin{subfigure}[b]{\textwidth}
         \centering
         \includegraphics[width=\textwidth]{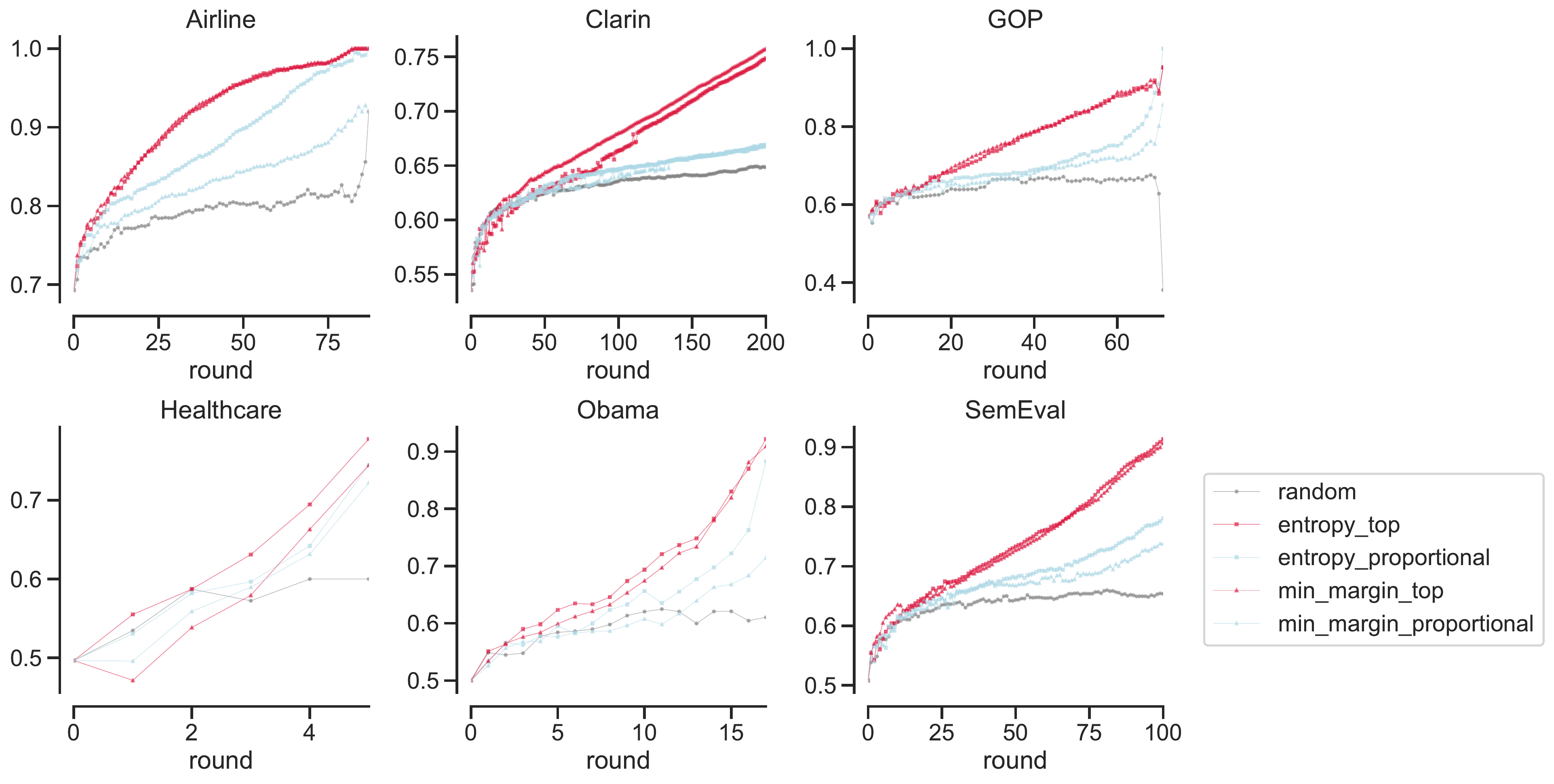}
         \caption{Sentiment classification}
         \label{fig:hitl_sentiment_unselected_results}
     \end{subfigure}
     \begin{subfigure}[b]{\textwidth}
         \centering
         \includegraphics[width=\textwidth]{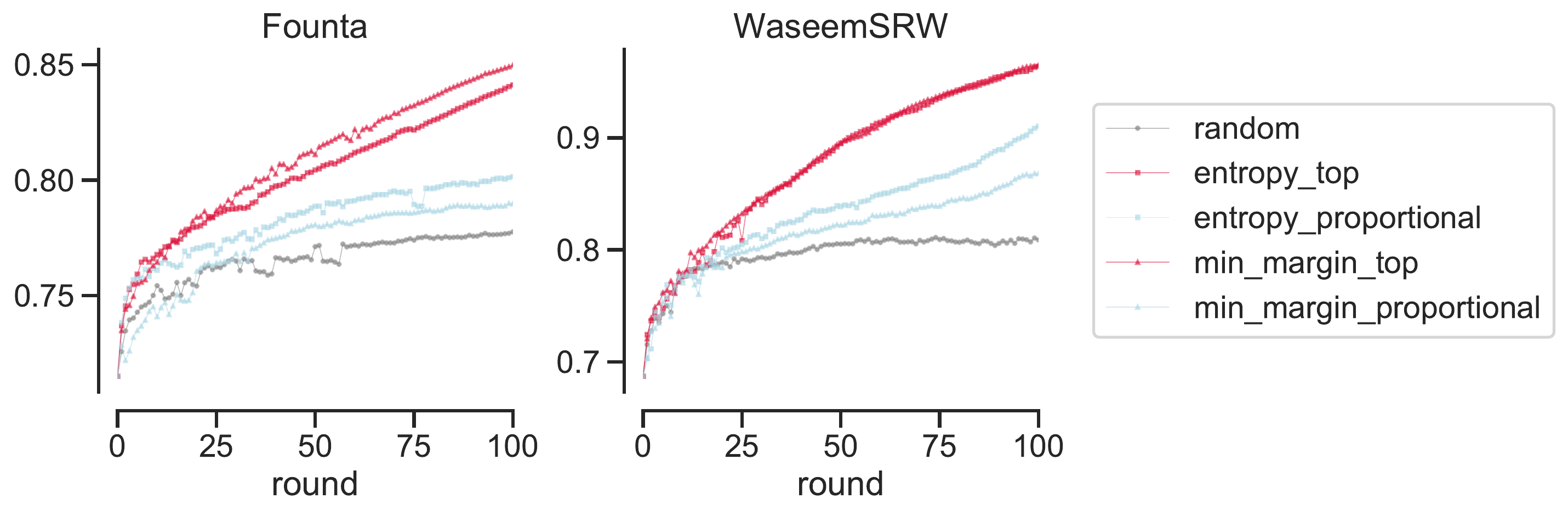}
         \caption{Abusive content detection}
         \label{fig:hitl_abusive_unselected_results}
     \end{subfigure}
     \begin{subfigure}[b]{\textwidth}
         \centering
         \includegraphics[width=\textwidth]{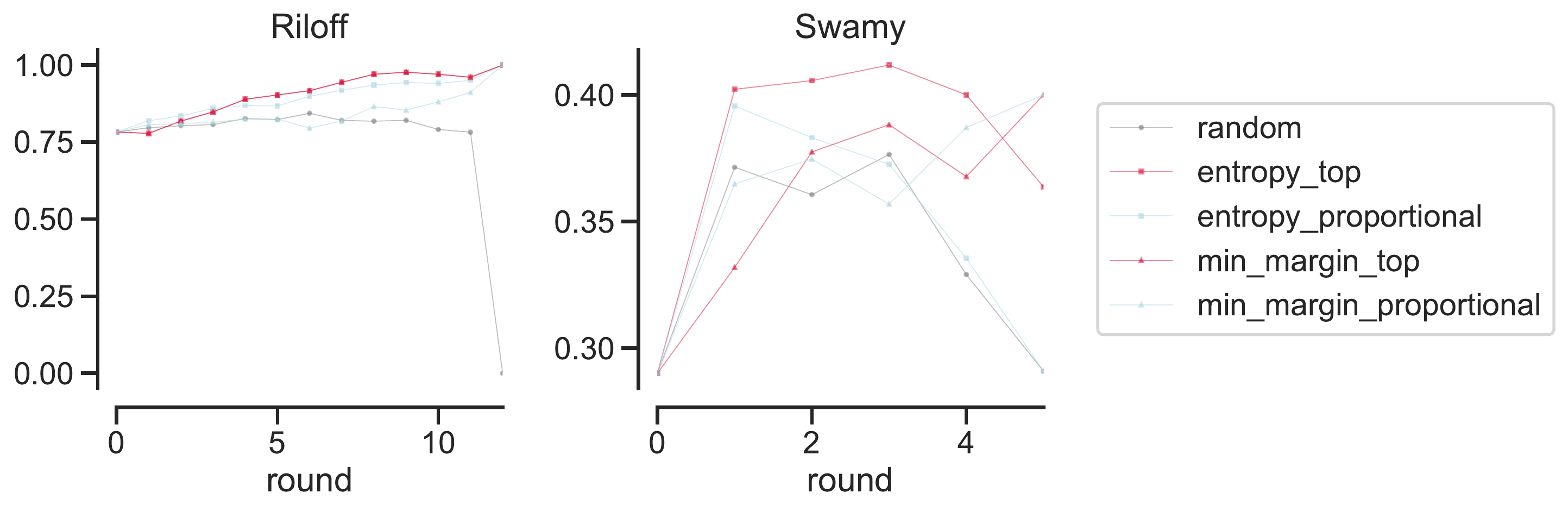}
         \caption{Uncertainty indicators}
         \label{fig:hitl_uncertainty_unselected_results}
     \end{subfigure}
    \caption[Active learning performance on unselected data across multiple classification tasks]{Progression of active learning classifier performance (micro f1-score) on the respective unselected data set across 100 rounds of active learning (200 for Clarin). The annotation budget for each round is 100 instances, and the model is warm started with 100 random samples of the training data. Data ordered alphabetically and X and Y axes are not shared.}
    \label{fig:hitl_all_classification_unselected_results}
\end{figure}

\paragraph{Evaluation on remaining dataset} Active learning systems often just track the test dataset performance. However, we observe another dataset which is not used for training, it is the left over dataset $X_{left}$ after selecting the examples in each round. $X_{left}$ is continously decreasing and tracking the performance of the model on $X_{left}$ can reveal how fast can an in-distribution dataset be accurately annotated using the specific querying strategy. This is suitable for simulation mode where the whole dataset ($X_{left} = X_{unlabeled}$) is already annotated.

\paragraph{Simulation Experiments} Human annotation for \pytail{} can be simulated.First, $X_train$ is set to $N=100$ random samples from $X_{unlabeled}$. In each round, $X_{select}$ is $K$ ($K$=100) instances from $X_{unlabeled}$ based on the scoring criterion described above. We conduct 100 rounds of active learning (200 for Clarin as it is a very large dataset) and evaluate the models using the micro-f1 score. We also compare against a model trained on the full data (Full).
The experimental results on the test split of each data are shown in figure \ref{fig:hitl_all_classification_test_results} and table \ref{tab:10pct-results}. We observe that the top $K$ strategy is usually the best followed by the proportional strategy across all data. For larger datasets we see that the model closes the gap very soon. 
We also show experimental results on the $X_{left}$ part of the training data in figure \ref{fig:hitl_all_classification_unselected_results}. We observe that the top $K$ strategy is consistently the best, followed by the proportional strategy across all data. The increase in performance on the $X_{left}$ is indicative of the fact that active learning ensures that the remaining data is actually easy to annotate without human correction. This evaluation presents a more practical usage pattern of ML models. This usage pattern requires annotating pre-selected and large $X_{unlabeled}$. In reality, once the dataset is selected, one is interested in reducing the size of $X_{train}$ to efficiently annotate the data. We think, it is in this setting that the active learning is most beneficial. If the user can achieve high labeling accuracy by annotating few samples, then the user's job is done.




\section{Conclusion}
We described experiments for evaluating active learning approaches for text classification tasks on tweet data. We introduced, \pytail{}, a user interface for active learning of NLP models by only requiring the user to update the labels for the model prediction if required. One limitation of our work is that our experiments are only conducted using simple linear model as they are easier to experiment with for sparse text features which we used for feature importance. However, the API does not place any restriction on the type of model. \pytail{} will publicly available as an open source tool.



\bibliography{references}
\bibliographystyle{plainnat}





\section*{Checklist}

The checklist follows the references.  Please
read the checklist guidelines carefully for information on how to answer these
questions.  For each question, change the default \answerTODO{} to \answerYes{},
\answerNo{}, or \answerNA{}.  You are strongly encouraged to include a {\bf
justification to your answer}, either by referencing the appropriate section of
your paper or providing a brief inline description.  For example:
\begin{itemize}
  \item Did you include the license to the code and datasets? \answerYes{See Section~\ref{gen_inst}.}
  \item Did you include the license to the code and datasets? \answerNo{The code and the data are proprietary.}
  \item Did you include the license to the code and datasets? \answerNA{}
\end{itemize}
Please do not modify the questions and only use the provided macros for your
answers.  Note that the Checklist section does not count towards the page
limit.  In your paper, please delete this instructions block and only keep the
Checklist section heading above along with the questions/answers below.

\begin{enumerate}

\item For all authors...
\begin{enumerate}
  \item Do the main claims made in the abstract and introduction accurately reflect the paper's contributions and scope?
    \answerYes{}
  \item Did you describe the limitations of your work?
    \answerYes{}
  \item Did you discuss any potential negative societal impacts of your work?
    \answerNA{}
  \item Have you read the ethics review guidelines and ensured that your paper conforms to them?
    \answerYes{}
\end{enumerate}

\item If you are including theoretical results...
\begin{enumerate}
  \item Did you state the full set of assumptions of all theoretical results?
    \answerNA{}
        \item Did you include complete proofs of all theoretical results?
    \answerNA{}
\end{enumerate}

\item If you ran experiments...
\begin{enumerate}
  \item Did you include the code, data, and instructions needed to reproduce the main experimental results (either in the supplemental material or as a URL)?
    \answerNo{No URL for anonymity. Our code is public.}
  \item Did you specify all the training details (e.g., data splits, hyperparameters, how they were chosen)?
    \answerYes{}
        \item Did you report error bars (e.g., with respect to the random seed after running experiments multiple times)?
    \answerNo{}
        \item Did you include the total amount of compute and the type of resources used (e.g., type of GPUs, internal cluster, or cloud provider)?
    \answerNo{No GPU used. All experiments on CPU.}
\end{enumerate}

\item If you are using existing assets (e.g., code, data, models) or curating/releasing new assets...
\begin{enumerate}
  \item If your work uses existing assets, did you cite the creators?
    \answerYes{}
  \item Did you mention the license of the assets?
    \answerNo{}
  \item Did you include any new assets either in the supplemental material or as a URL?
    \answerNo{}
  \item Did you discuss whether and how consent was obtained from people whose data you're using/curating?
    \answerNA{}
  \item Did you discuss whether the data you are using/curating contains personally identifiable information or offensive content?
    \answerYes{}
\end{enumerate}

\item If you used crowdsourcing or conducted research with human subjects...
\begin{enumerate}
  \item Did you include the full text of instructions given to participants and screenshots, if applicable?
    \answerNA{}
  \item Did you describe any potential participant risks, with links to Institutional Review Board (IRB) approvals, if applicable?
    \answerNA{}
  \item Did you include the estimated hourly wage paid to participants and the total amount spent on participant compensation?
    \answerNA{}
\end{enumerate}

\end{enumerate}


\appendix

\end{document}